\documentclass[runningheads]{llncs}
\usepackage{graphicx}
\usepackage{amsmath,amssymb} 
\usepackage{color}
\usepackage[width=122mm,left=12mm,paperwidth=146mm,height=193mm,top=12mm,paperheight=217mm]{geometry}

\usepackage{bm}
\usepackage{tabularx}
\usepackage{hyperref}
\usepackage{caption}

\begin{document}
\pagestyle{headings}
\mainmatter

\title{Deep Discriminative Model for Video Classification} 

\titlerunning{Deep Discriminative Model}

\authorrunning{M. Tavakolian and A. Hadid}

\author{Mohammad Tavakolian and Abdenour Hadid}


\institute{Center for Machine Vision and Signal Analysis (CMVS)\\
	University of Oulu, Finland\\
	\email{ firstname.lastname@oulu.fi}
}

\maketitle

\begin{abstract}
This paper presents a new deep learning approach for video-based scene classification. We design a Heterogeneous Deep Discriminative Model (HDDM) whose parameters are initialized by performing an unsupervised pre-training in a layer-wise fashion using Gaussian Restricted Boltzmann Machines (GRBM). In order to avoid the redundancy of adjacent frames, we extract spatiotemporal variation patterns within frames and represent them sparsely using Sparse Cubic Symmetrical Pattern (SCSP). Then, a pre-initialized HDDM is separately trained using the videos of each class to learn class-specific models. According to the minimum reconstruction error from the learnt class-specific models, a weighted voting strategy is employed for the classification. The performance of the proposed method is extensively evaluated on two action recognition datasets; UCF101 and Hollywood II, and three dynamic texture and dynamic scene datasets; DynTex, YUPENN, and Maryland. The experimental results and comparisons against state-of-the-art methods demonstrate that the proposed method consistently achieves superior performance on all datasets. 

\end{abstract}

\vspace{-2.3em}
\section{Introduction}
\vspace{-0.5em}
Through the recent surge in digital content, video data has become an indisputable part of today's life. This has stimulated the evolution of advanced approaches for a wide range of video understanding applications. In this context, the understanding and classification of video content have gained a substantial research interest among the computer vision community. However, the automatic classification of scene in videos is subject to a number of challenges, including a range of natural variations in short videos such as illumination variations, viewpoint changes, and camera motions. Moreover, scene classification differs from the conventional object detection or classification, because a scene is composed of several entities which are often organized in a random layout. Therefore, devising an accurate, efficient and robust representation of videos is essential to deal with these challenges.  

To achieve an effective representation of a scene in videos, we can model videos' spatiotemporal motion patterns using the concept of dynamic textures. Videos comprise dynamic textures which inherently exhibit spatial and temporal regularities of a scene or an object. Dynamic textures widely exist in real-world video data, e.g. regular rigid motion like windmill, chaotic motion such as smoke and water turbulences, and sophisticated motion caused by camera panning and zooming. The modeling of dynamic textures in videos is challenging but very important for computer vision applications such as video classification, dynamic texture synthesis, and motion segmentation. 

Despite all challenges, great efforts have been devoted to find a robust and powerful solution for video-based scene classification tasks. Furthermore, it has been commonly substantiated that an effective representation of the video content is a crucial step towards resolving the problem of dynamic texture classification. In previous years, a substantial number of approaches for video representation have been proposed, e.g. Linear Dynamic System (LDS) based methods \cite{c1}, Local Binary Pattern (LBP) based methods \cite{c2}, and Wavelet-based methods \cite{c3}. Unfortunately, the current methods are sensitive to a wide range of variations such as viewpoint changes, object deformations, and illumination variations. Coupled with these drawbacks, other methods frequently model the video information within consecutive frames on a geometric surface  represented by a subspace \cite{c4}, a combination of subspaces \cite{c5}, a point on the Grassmann manifold \cite{c6}, or Lie Group of Riemannian manifold \cite{c7}. These require prior assumptions regarding specific category of the geometric surface on which samples of the video are assumed to lie. 

\begin{figure}[t]
\begin{center}
   \includegraphics[width=0.55\linewidth]{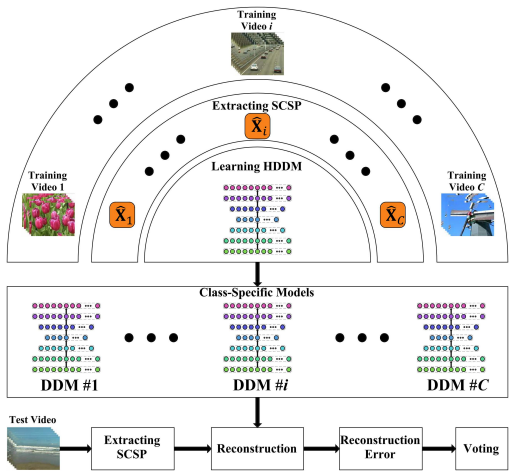}
\end{center}
   \caption{The block diagram of the proposed DDM framework for video classification.}
\label{Fig1}
\vspace{-1.5em}
\end{figure}

On the other hand, deep learning has recently achieved significant success in a number of areas \cite{c8,c9,c10}, including video scene classification \cite{c11,c12,c13,c14}. Unlike the conventional methods, which fail to model discontinuous rigid motions, deep learning based approaches have a great modeling capacity and can learn discriminative representations in videos. However, the current techniques have mostly been devised to deal with fixed-length video sequences. They fail to deal with long sequences due to their limited temporal coverage. This paper presents a novel deep learning approach, which does not assume any biased knowledge about the concept of data and it automatically explores the structure of the complex non-linear surface on which the samples of the video are present. According to the block diagram in Figure \ref{Fig1}, our proposed method defines a Heterogeneous Deep Discriminative Model (HDDM) whose weights are initialized by an unsupervised layer-wise pre-training stage using Gaussian Restricted Boltzmann Machines (GRBM) \cite{c54}. The initialized HDDM is then separately trained for each class using all videos of that class in order to learn a Deep Discriminative Model (DDM) for every class. The training is done so that the DDM learns to specifically represent videos of that class. Therefore, a class specific model is made to learn the structure and the geometry of the complex non-linear surface on which video sequences of that class exist. Also, we represent the raw video data using Sparse Cubic Symmetrical Pattern (SCSP) to capture long-range spatiotemporal patterns and reduce the redundancy between adjacent frames. For the classification of a given query video, we first represent the video based on the learnt class-specific DDMs. The representation errors from the respective DDMs are then computed and a weighted voting strategy is used to assign a class label to the query video. 

The main contributions of our proposed Deep Discriminative Model (DDM) are the followings. First, a novel deep learning based framework is introduced for video classification (Sec. 3). Moreover, we develop a Sparse Cubic Symmetrical Pattern (SCSP) to avoid the redundancy in video sequences and reduce the computational cost, and a weighted voting strategies is utilized for classification (Sec. 4). Finally, extensive experiments are conducted along with comparisons against state-of-the-art methods for video classification. The experimental results demonstrate that the proposed method achieves superior performance compared to the state-of-the-art methods (Sec. 5).

\vspace{-1em}
\section{Related Work}
\vspace{-0.5em}
Several approaches have been proposed for video classification \cite{c3,c15,c16}. A popular approach is Linear Dynamic System (LDS) \cite{c1,c15}, which is known as a probabilistic generative model defined over space and time. LDS approximates hidden states using Principal Component Analysis (PCA), and describes their trajectory as time evolves. LDS has obvious drawbacks due to its sensitivity to external variations. In order to overcome this limitation, Closed-Loop LDS (CLDS) \cite{c17} was proposed. However, CLDS tends to fail to capture some discontinuous rigid motions due to its simplistic linearity. Local Binary Pattern (LBP) based methods \cite{c2} have been widely used in texture analysis. Zhao \textit{et al.} \cite{c18} extended LBP to the space and the time domains and proposed two LBP variants: 1) Volume Local Binary Pattern (VLBP) \cite{c18} which combines both the spatial and the temporal variations of the video; 2) Local Binary Pattern on Three Orthogonal Planes (LBP-TOP) \cite{c18}, which computes LBP in three individual $x-y$, $x-t$, and $y-t$ planes to describe the video. Likewise, other versions of LBP-TOP, such as Local Ternary Pattern on Three Orthogonal Planes (LTP-TOP) \cite{c19} and Local Phase Quantization on Three Orthogonal Planes (LPQ-TOP) \cite{c19}, have been proposed. Although they are all effective in capturing the spatiotemporal information, they rarely achieve a satisfactory performance in the presence of camera motions.

Recently, there is a huge growing research interest in deep learning methods in various areas of computer vision, beating the state-of-the-art techniques \cite{c9,c11,c12,c13,c14}. Deep learning methods set up numerous recognition records in image classification \cite{c20}, object detection \cite{c21}, face recognition and verification \cite{c22}, and image set classification \cite{c10}. Deep models, such as Deep Belief Networks and stacked autoencoders, have much more expressive power than traditional shallow models and can be effectively trained with layer-wise pre-training and fine-tuning \cite{c23}. Stacked autoencoders have been successfully used for feature extraction \cite{c24}. Also, they can be used to model complex relationships between variables due to the composition of several levels of non-linearity \cite{c24}. Xie \textit{et al.} \cite{c25} modeled relationships between noisy and clean images using stacked denoising autoencoders. Although, deep autoencoders are rarely used to model time series data, there are researches on using variants of Restricted Boltzmann Machine (RBM) \cite{c26} for specific time series data such as human motion \cite{c27}.  On the other hand, some convolutional architectures have been used to learn spatiotemporal features from video data \cite{c28}. Kaparthy \textit{et al.} \cite{c11} used a deep structure of Convolutional Neural Networks (CNN) and tested it on a large scale video dataset. By learning long range motion features via training a hierarchy of multiple convolutional layers, they showed that their framework is just marginally better than single frame-based methods.  Simonyan \textit{et al.} \cite{c12} designed Two-Stream CNN which includes the spatial and the temporal networks. They took advantage of ImageNet dataset for pre-training and calculated the optical flow to explicitly capture the motion information. Tran \textit{et al.} \cite{c13} investigated 3D CNNs \cite{c29} on realistic (captured in the wild) and large-scale video datasets. They tried to learn both the spatial and temporal features with 3D convolution operations. Sun \textit{et al.} \cite{c14} proposed a factorized spatiotemporal CNN and exploited different ways to decompose 3D convolutional kernels. 

The long-range temporal structure plays an important role in understanding the dynamics of events in videos. However, mainstream CNN frameworks usually focus on appearances and short-term motions. Thus, they lack the capacity to incorporate the long-range temporal structure. Recently, few other attempts (mostly relying on dense temporal sampling with a pre-defined sampling interval) have been proposed to deal with this problem \cite{c30,c31}. This approach would incur excessive computational cost and is not applicable to real-world long video sequences. It also poses the risk of missing important information for videos that are longer than the maximal sequence length. Our proposed method deals with this problem by extracting Sparse Cubic Symmetrical Patterns (SCSP) from video sequences to feed its autoencoder structure (Sec. 4.1). In terms of spatiotemporal structure modeling, a key observation is that consecutive frames are highly redundant. Therefore, dense temporal sampling, which results in highly similar sampled frames, is unnecessary. Instead, a sparse spatiotemporal representation will be more favorable in this case. Also, autoencoders reduce the dimension and keep as much important information as possible, and remove noise. Furthermore, combining them with RBMs helps the model to learn more complicated video structures based on their non-linearity.


\section{The Proposed Deep Discriminative Model}
\begin{figure}[t]
\begin{center}
   \includegraphics[width=0.9\linewidth]{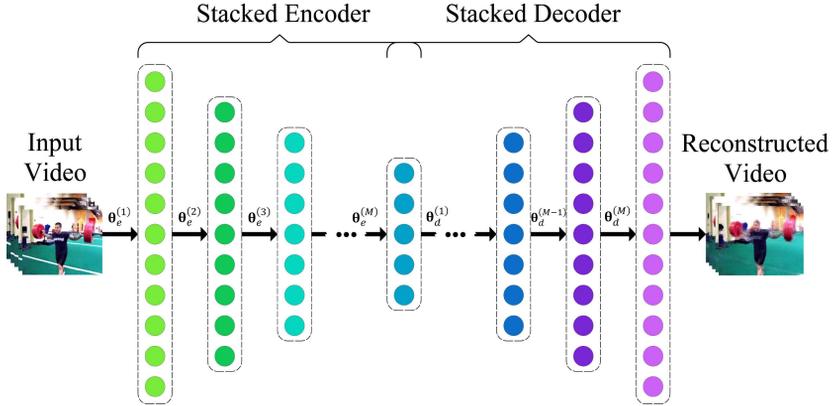}
\end{center}
\vspace{-1.2em}
   \caption{The configuration of the proposed Heterogeneous Deep Discriminative Model.}
\label{Fig2}
\vspace{-1.5em}
\end{figure}

\vspace{-0.7em}
We first define a Heterogeneous Deep Discriminative Model (HDDM) which will be used to learn the underlying structure of the data in Sec 4.2. The architecture of the HDDM is shown in Figure \ref{Fig2}. Generally, an appropriate parameter initialization is inevitable for deep neural networks to have a satisfactory performance. Therefore, we initialize the parameters of HDDM by performing pre-training in a greedy layer-wise framework using Gaussian Restricted Boltzmann Machines. The initialized HDDM is separately fine-tuned for each of the $C$ classes of the training videos. Therefore, we end up with a total of $C$ fine-tuned Deep Discriminative Models (DDMs). Then, the fined-tuned models are used for video classification. 
\vspace{-1em}
\subsection{The Heterogeneous Deep Discriminative Model}
\vspace{-0.5em}
As can be seen in Figure \ref{Fig2}, the proposed HDDM is based on an autoencoder which comprises multiple encoder and decoder layers. In the proposed autoencoder structure, both the encoder and the decoder have $M$ hidden layers each such that the $M$-th layer of the encoder is considered as the first layer of the decoder. The encoder section represents the input data in a lower dimension. The encoder consists of a combination of non-linear functions $s\left( \cdot  \right)$ used to map the input data $\mathbf{x}$ to a representation $\mathbf{h}$ given by
\vspace{-0.5em}
\begin{equation}
\mathbf{h} = s\left(\mathbf{x} \Big | \bm{\theta} _{e}^{(1)}, \bm{\theta} _{e}^{(2)}, \dots, \bm{\theta} _{e}^{(M)}  \right)
\vspace{-0.7em}
\end{equation}
where $\bm{\theta} _{e}^{(i)}=\left\{ \mathbf{W}_{e}^{(i)},\mathbf{b}_{e}^{(i)} \right\}$ denotes the parameters of the $i$-the encoder layer. So, $\mathbf{W}_{e}^{(i)}\in {\mathbb{R}^{{{n}_{i}}\times {{n}_{i-1}}}}$ is the encoder weight matrix for layer $i$ having $n_i$ nodes, $\mathbf{b}_{e}^{(i)}\in {\mathbb{R}^{{{n}_{i}}}}$ is the bias vector and $s\left( \cdot  \right)$ is a non-linear sigmoid activation function.  The encoder parameters are learnt by combining the encoder with the decoder and jointly train the encoder-decoder structure to represent the input data by optimizing a cost function. Hence, the decoder can be defined as series of non-linear functions, which calculate an approximation of the input $\mathbf{x}$ from the encoder output $\mathbf{h}$. The approximated output $\tilde{\mathbf{x}}$ of the decoder is obtained by
\begin{equation}
\vspace{-0.7em}
\tilde{\mathbf{x}} = s\left(\mathbf{h} \Big | \bm{\theta} _{d}^{(1)}, \bm{\theta} _{d}^{(2)}, \dots, \bm{\theta} _{d}^{(M)}  \right)
\vspace{-0.3em}
\end{equation}
where $\bm{\theta} _{d}^{(j)}=\left\{ \mathbf{W}_{d}^{(j)},\mathbf{b}_{d}^{(j)} \right\}$ are the parameters of the $j$-the decoder layer. Consequently, we represent the complete encoder-decoder structure by its parameters ${\bm{\theta }_{HDDM}}=\left\{ {{\bm{\theta} }_\mathbf{W}},{\bm{\theta }_\mathbf{b}} \right\}$, where ${\bm{\theta }_\mathbf{W}}=\left\{ \mathbf{W}_{e}^{(i)},\mathbf{W}_{d}^{(i)} \right\}_{i=1}^{M}$ and ${\bm{\theta }_\mathbf{b}}=\left\{ \mathbf{b}_{e}^{(i)},\mathbf{b}_{d}^{(i)} \right\}_{i=1}^{M}$. 

\vspace{-0.7em}
\subsection{Parameter Initialization}

We train the defined HDDM with videos of each class individually, which results in class-specific models. The training is performed through stochastic gradient descent with back propagation \cite{c32}. The training may not yield to desirable results if the HDDM is initialized with inappropriate weights. Thus, the parameters of the model are first initialized through an unsupervised pre-training phase. For this purpose, a greedy layer-wise approach is adopted and Gaussian RBMs \cite{c54} are used. 

Basically, a standard RBM \cite{c26} is used for binary stochastic data. We therefore use an extension of RBM to process real valued data by appropriate modifications in its energy function. Gaussian RBM (GRBM) \cite{c54} is one such popular extension whose energy function is defined by changing the bias term of the visible layer. 
\vspace{-1em}
\begin{equation}
{{E}_{GRBM}}\left( v,h \right)=\sum\limits_{i}{\frac{{{\left( {{v}_{i}}-{{b}_{i}} \right)}^{2}}}{2\sigma _{i}^{2}}}-\sum\limits_{j}{{{c}_{j}}{{h}_{j}}-} \sum\limits_{ij}{{{w}_{ij}}\frac{{{v}_{i}}}{{{\sigma }_{i}}}{{h}_{j}}}
\vspace{-0.7em}
\end{equation}
where $\mathbf{W}$ is the weight matrix, and $\mathbf{b}$ and $\mathbf{c}$ are the biases of the visible and the hidden layer, respectively. We use a numerical technique called Contrastive Divergence (CD) \cite{c33} to learn the model parameter $\left\{ \mathbf{W},\mathbf{b},\mathbf{c} \right\}$ of the GRBM in the training phase. $v_i$ and $h_j$ denote the visible layer and the hidden layer’s nodes, respectively. Also, ${{\sigma }_{i}}$ is the standard deviation of the real valued Gaussian distributed inputs to the visible node $v_i$. It is possible to learn ${{\sigma }_{i}}$ for each visible unit but it becomes arduous when using CD for GRBM parameter learning. We therefore use another approach and set ${{\sigma }_{i}}$ to a constant value.

Since there are no intra-layer node connections, result derivation becomes easily manageable for the RBM to the contrary of most directed graphical models. The probability distributions for GRBM are given by
\begin{align}
\rho \left( {{h}_{j}}\left| \mathbf{v} \right. \right) &= s\left( \sum\nolimits_{i}{{{w}_{ij}}{{v}_{i}}+{{c}_{j}}} \right) \\
\rho \left( {{v}_{i}}\left| \mathbf{h} \right. \right) &= \frac{1}{{{\sigma }_{i}}\sqrt{2\pi }}\exp \left( \frac{-{{\left( {{v}_{i}}-{{u}_{i}} \right)}^{2}}}{2\sigma _{i}^{2}} \right)
\end{align}
where
\begin{equation}
{{u}_{i}} = {{b}_{i}}+\sigma _{i}^{2}\sum\limits_{i}{{{w}_{ij}}{{h}_{j}}}
\end{equation}

Since our data are real-valued, we use GRBMs to initialize the parameters of the proposed HDDM. In this case, we consider two stacked layers at a time to obtain the GRBM parameters during the learning process. First, the input layer nodes and the first hidden layer nodes are considered as the visible units $v$ and the hidden unit $h$ of the first GRBM, respectively, and their respective parameters are obtained. The activations of the first GRBM’s hidden units are then used as an input to train the second GRBM. We repeat this process for all four encoder hidden layers. The weights learnt for the encoder layers are then tied to the corresponding decoder layers. 

\vspace{-1em}
\section{Video Classification Procedure}

In this section, we describe how to classify query videos using the representation error. Assume that there are $C$ training videos $\left\{ {\mathbf{X}_{c}} \right\}_{c=1}^{C}$ with the corresponding class labels ${{y}_{c}}\in \left\{ 1,2,\cdots ,C \right\}$. A video sequence is denoted by ${\mathbf{X}_{c}}=\left\{ {\mathbf{x}^{(t)}} \right\}_{t=1}^{T}$, where $\mathbf{x}^{(t)}$ contains raw pixel values of the frame at time $t$. The problem is to assign class $y_q$ to the query video $\mathbf{X}_q$.
\vspace{-1em}
\subsection{Sparse Cubic Symmetrical Patterns}

We represent dynamic textures by video blocks, video volumes spanning over both the spatial and temporal domains, to jointly model the spatial and temporal information. Since there are strong correlations between adjacent regions of scenes (which cause redundancy), we devise an approach to extract a sparse representation of the spatiotemporal encoded features. As a result, the less important information is discarded which makes the deep discriminative model representation more efficacious. For this purpose, we design a volumetric based descriptor to capture the spatiotemporal variations in the scene. Given a video, we first decompose it into a batch of small cubic spatiotemporal volumes. We only consider the video cubes of small size ($w \times h \times d$ pixels), which consists of relatively simple content that can be generated with few components.

\begin{figure}[t]
\begin{center}
   \includegraphics[width=0.9\linewidth]{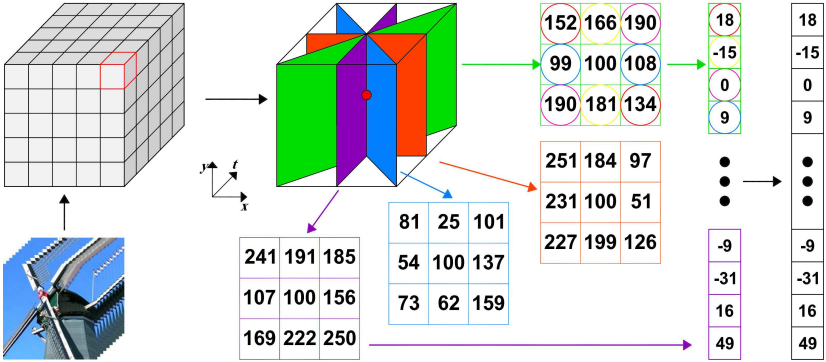}
\end{center}
   \caption{An example of extracting symmetric signed magnitude variation pattern in a volumetric block of the video sequence.}
\label{Fig3}
\vspace{-1.5em}
\end{figure}

Figure \ref{Fig3} illustrates the feature extraction process. We divide each series of frames $\mathbf{X}=\left\{ {\mathbf{x}^{(t)}} \right\}_{t=1}^{T}$ into $w \times h \times d$ distinct non-overlapping uniformly spaced cubic blocks and extract symmetric spatiotemporal variation pattern for every block which results in a feature vector of the corresponding block. Consequently, each video sequence $\mathbf{X}$ is encoded in terms of the symmetric signed magnitude variation patterns, denoted as ${\mathbf{x}_{E}}\in {\mathbb{R}^{d}}$, obtained by concatenating the feature vectors of all cubic blocks spanning over the entire video sequence. We do not consider the last one or two frames of video sequence if they do not fit in the cubic block structure. This does not affect the algorithm’s performance due to the correlation between consecutive frames. 

Given any pixel $x_{o}^{(t)}$, we represent the neighboring pixels by $x_{1}^{(t)},\cdots ,x_{P}^{(t)}$. The symmetric spatiotemporal variations for $j$-th plane is calculated as 
\begin{equation}
{{F}_{j}}\left( x_{o}^{(t)} \right)=\biguplus _{p=1}^{\frac{P}{2}}\left( x_{p}^{(t)}-x_{p+\frac{P}{2}}^{(t)} \right)
\end{equation}
where $x_{p}^{(t)}$ and $x_{p+\frac{P}{2}}^{(t)}$ are two symmetric neighbors of pixel $x_{o}^{(t)}$. Also, $\biguplus$ denotes the concatenation operator.

The aforementioned feature vectors are organized into the columns of a matrix $\mathbf{D}\in {\mathbb{R}^{d\times N}}$, where $d$ is the dimension of the feature vectors and $N$ is the total number of videos. In variable-length videos, we temporally partition the video sequences into non-overlapping segments with a fixed length $k$ and extract features from the cubic blocks within each segment, separately. Then, we place the extracted features of each section into a column of the matrix $\mathbf{D}$. Here, we call matrix $\mathbf{D}$ the dictionary and aim to find a sparse representation $\hat{\mathbf{x}}$ of the encoded video $\mathbf{x}_E$, ${\mathbf{x}_{E}}=\mathbf{D}\hat{\mathbf{x}}$ by basis matching pursuit \cite{c34} such that 
\begin{equation}
\underset{{\overset{\scriptscriptstyle\frown}{\mathbf{x}}}}{\mathop{\min }}\,\frac{1}{2}\left\| {\mathbf{x}_{E}}-\mathbf{D}\hat{\mathbf{x}} \right\|_{2}^{2}+\lambda {{\left\| {\hat{\mathbf{x}}} \right\|}_{1}}
\end{equation}
where $\lambda$ is a slack variable and ${{\left\| \cdot  \right\|}_{1}}$ is the sparsity including ${{\ell }_{1}}$ norm. The slack variable balances the trade-off between fitting data perfectly and employing a sparse solution. For further improvements, we represent each color channel, individually. Also, we reshape the sparsely represented vector $\hat{\mathbf{x}}$ into a 3D structure of $\hat{\mathbf{X}}=\left\{ {{{\hat{\mathbf{x}}}^{(l)}}} \right\}_{l=1}^{L}$, where $L$ is the length of the structure. We feed the proposed deep model with Sparse Cubic Symmetrical Patterns (SCSP) instead of raw videos.

For notational simplicity, we will consider the sparsely represented $\hat{\mathbf{X}}=\left\{ {{{\hat{\mathbf{x}}}^{(l)}}} \right\}_{l=1}^{L}$ as a sequence of frames with length $L$ and denote it by ${\mathbf{X}}=\left\{ {{{{\mathbf{x}}}^{(l)}}} \right\}_{l=1}^{L}$ hereafter.

\subsection{Learning DDMs of the Training Classes}

In order to initialize the parameters of the HDDM using GRBMs, we randomly shuffle a small subset, containing video sequences from all classes (of the training video sequences). We use this subset for layer-wise GRBM training of all encoder layers. The parameters of the decoder layers are then configured with their corresponding tied parameters of the encoder layers. This process assures us that rarely does the proposed network gets stuck in a local minimum.

We define a cost function based on the representation error over all frames of the video for learning class-specific parameters. In order to avoid over-fitting and enhance the generalization of the learnt model to unseen test data, the regularization terms are added to the cost function of the HDDM. A weight decay penalty term $J_{wd}$ and a sparsity constraint $J_{sp}$ are added.
\begin{equation}
{{J}_{reg}}\left( {{\theta }_{HDDM}}\left| {{x}^{(l)}}\in {{X}_{c}} \right. \right)=\sum{{{\left\| {{x}^{(l)}}-{{{\tilde{x}}}^{(l)}} \right\|}^{2}}}+{{\lambda }_{wd}}{{J}_{wd}}+{{\lambda }_{sp}}{{J}_{sp}}
\end{equation}
where ${\lambda }_{wd}$ and ${\lambda }_{sp}$ are regularization parameters. ${J}_{wd}$ guarantees small values of weights for all hidden units and ensures that dropping out will not happen for hidden layers’ units. It is defined as the summation of the Frobenius norm of all weight matrices:
\begin{equation}
{{J}_{wd}}=\sum\limits_{i=1}^{M}{\left\| \mathbf{W}_{e}^{(i)} \right\|_{F}^{2}}+\sum\limits_{i=1}^{M}{\left\| \mathbf{W}_{d}^{(i)} \right\|_{F}^{2}}
\end{equation}

Moreover, $J_{sp}$ enforces that the mean activation $\bar{\rho }_{j}^{(i)}$ (over all training samples) of the $j$-th unit of the $i$-th hidden layer is as close as possible to the sparsity target $\rho$ which is a very small constant. $J_{sp}$ is further defined based on the KL divergence as 
\begin{equation}
{{J}_{sp}}=\sum\limits_{i=1}^{2M-1}{\sum\limits_{j}{\rho \log \frac{\rho }{\bar{\rho }_{j}^{(i)}}+\left( 1-\rho  \right)\log \frac{1-\rho }{1-\bar{\rho }_{j}^{(i)}}}}
\end{equation}

Therefore, a class specific model $\bm{\theta}_c$ is obtained by optimizing the regularized cost function $J_{reg}$ over all frames of the class $\mathbf{X}_c$. 
\begin{equation}
{\bm{\theta }_{c}}=\arg \underset{{\bm{\theta }_{HDDM}}}{\mathop{\min }}\,{{J}_{reg}}\left( {\bm{\theta }_{HDDM}}\left| {\mathbf{x}^{(l)}}\in {\mathbf{X}_{C}} \right. \right)
\end{equation}

We note that our proposed model is easily scalable. Enrolling new classes would not require re-training on the complete database. Instead, the class-specific models for the added classes can be learnt independently of the existing classes.

\subsection{Classification}

Given a query video sequence ${\mathbf{X}_{q}}=\left\{ {\mathbf{x}^{(t)}} \right\}_{t=1}^{{{T}_{q}}}$, we first extract SCSPs and then separately reconstruct them using all class-specific DDMs $\bm{\theta}_{c}$, $c=1,\cdots ,C$, using Equations (1) and (2). Suppose $\tilde{\mathbf{x}}_{c}^{\left( l \right)}$ is the $l$-th frame of the reconstructed query video sequence ${{\tilde{\mathbf{X}}_{{{q}_{c}}}}}$ based on the $c$-th class model $\bm{\theta}_{c}$. We obtain the reconstruction errors, i.e. ${{\left\| {{x}^{(l)}}-\tilde{x}_{c}^{(l)} \right\|}_{2}}$, from all class specific models; then, a weighted voting strategy is employed to determine the class label of the given query video sequence $\mathbf{X}_q$. Each query video sequence’s frame $\mathbf{x}^{(l)}$ casts a vote to all classes. Using the reconstruction error of each class’s model, we assign a weight to the casted vote to each class. The weight ${{\mu }_{c}}\left( {{x}^{(l)}} \right)$ of the vote casted by a frame $\mathbf{x}^{(l)}$ to class $c$ is defined as
\begin{equation}
{{\mu }_{c}}\left( {\mathbf{x}^{(l)}} \right)=\exp \left( -{{\left\| {\mathbf{x}^{(l)}}-\tilde{\mathbf{x}}_{c}^{(l)} \right\|}_{2}} \right)
\end{equation}

The candidate class which achieves the highest accumulated weight from all frames of $\mathbf{X}_q$ is declared as the class $y_q$ of the query video sequence:
\begin{equation}
{{y}_{q}}=\arg \underset{c}{\mathop{\max }}\,\sum\limits_{{\mathbf{X}_{q}}}{{{\mu }_{c}}\left( {\mathbf{x}^{\left( l \right)}} \right)}
\end{equation}

\vspace{-1.5em}
\section{Experimental Analysis}
\vspace{-1em}
We extensively evaluate the performance of the proposed method on five benchmarking datasets  including UCF101 \cite{c35} and Hollywood II \cite{c36} datasets for action recognition, DynTex dataset \cite{c16} for dynamic texture recognition, and YUPENN \cite{c37} and Maryland \cite{c38} datasets for dynamic scene classification task.
\vspace{-2em}
\subsection{Parameter Setting}
\vspace{-0.5em}
We performed a grid search to obtain the optimal parameters and conducted experiments on a validation set. To be specific, the initial weights for layer-wise GRBM training are drawn from a uniform random distribution in the range of $\left[-0.005, 0.005\right]$. Contrastive Divergence (CD) was used to train GRBMs on 200 randomly selected videos from the training data. Mini-batches of 32 videos were used and the training was done for 50 epochs. A fixed learning rate of $10^{-3}$ was used. To train the pre-initialized HDDM to learn class-specific models, we used an annealed learning rate that is started with $2\times 10^{-3}$ and multiplied by a factor of $0.6$ per epoch. We chose $\ell_2$ weight decay ($\lambda_{wd}$) to be $0.01$, a sparsity target ($\rho$) of $0.001$, and non-sparsity penalty term ($\lambda_{sp}$) of 0.5. The training was performed by considering a mini-batch of 10 videos for 20 epochs. 

The size of volumetric blocks in SCSP also affects the performance of the algorithm. Therefore, we conducted an empirical study on different sizes of video blocks in Table \ref{BlockSize}. It is observed from Table \ref{BlockSize} that the best result is achieved with the block size of $3 \times 3 \times 3$. With very small blocks (e.g. $1 \times 1 \times 3$), few spatiotemporal regions are captured and the model will have problems on dealing with the scene variations. Moreover, the blocks of large sizes (e.g. $7 \times 7 \times 5$) carry too much information that does not improve the model's performance.

In order to determine the number of layers and the number of units in each layer, we employed a multi-resolution search strategy. The ideas is to test some values from a large parameter range, choose a few best configurations, and then test again with smaller steps around these values. We tested the model with the escalating number of layers \cite{Larochelle2007} and stopped where the performance reaches the highest rate on the validation set. The hidden layers sizes varies in the range of [250, 1000]. 

\vspace{-0.8cm}
\begin{table}
\centering
\scriptsize{
\caption{Comparison of the proposed method's accuracy (\%) on the UCF101 database \cite{c35} with different block sizes for SCSP.}
\begin{tabular}{|l|c|c|c|c|c|c|c|}
\hline
\textbf{Block Size} & $1\times 1\times 3$ & $1\times 1\times 5$ & $3\times 3\times 3$ & $3\times 3\times 5$ & $5\times 5\times 3$ & $5\times 5\times 5$ & $7\times 7\times 5$  \\ \hline
\textbf{Accuracy} & 87.3 & 91.2 & 94.3 & 92.5 & 89.4 & 84.3 & 79.5  \\ \hline
\end{tabular}
\label{BlockSize}}
\vspace{-10ex}
\end{table}

\subsection{Human Action Recognition}

We conducted experiments on two benchmark action recognition datasets, i.e. UCF101 \cite{c35} and Hollywood II \cite{c36} datasets, and compared the performance of the proposed method against state-of-the-art approaches.

\textit{The UCF101 dataset} \cite{c35} is composed of realistic web videos which are typically captured with large variations in camera motion, object appearance/scale, viewpoint, cluttered background, and illumination variations. It has 101 categories of human actions ranging from daily life to sports. The UCF101 contains 13,320 videos with an average length of 180 frames. It has three splits setting to separate the dataset into training and testing videos. We report the average classification accuracy over these three splits.

We compare the average accuracy performance of our proposed DDM with both the traditional and deep learning-based benchmark methods for human action recognition in Table \ref{Table1}. Our model obtains an average accuracy of $91.5\%$. However, the accuracy of DDM on the UCF101 is less than that of KVMF \cite{c48} by $1.6\%$. We argue that the performance of DDM degrades since it only captures short range spatiotemporal information in the video sequence. The videos in UCF101 exhibit significant temporal variations. Moreover, the severe camera movements increase the complexity of video's dynamics and make data reconstruction challenging. These issues bring up difficulties for the algorithm to focus on the action happening at each time instance.

To tackle this problem, we feed the extracted SCSP features to our DDM. The proposed SCSP extracts detailed spatiotemporal information by capturing the spatiotemporal variations of the video sequence within small volumetric blocks. By representing this information sparsely, it not only covers the whole length of the video sequence, but also decreases the redundancy of data. In this way, SCSP increases the discriminability of samples in the feature space in which the similar samples are mapped close to each other and dissimilar ones are mapped far apart. Therefore, the DDM can readily learn the underlying structure of each class. As can be seen from Table \ref{Table1}, the performance of our DDM improves by using SCSP features.
\begin{table}
\vspace{-2.35em}
\centering
\scriptsize{
\caption{Comparison of the average classification accuracy of DDM against state-of-the-art methods on the UCF101 dataset \cite{c35}.}
\begin{tabularx}{\columnwidth}{@{\extracolsep{\fill}}|p{0.45\columnwidth}p{0.5\columnwidth}|}
\hline
\multicolumn{1}{|c}{\textbf{Method}} & \multicolumn{1}{c|}{\textbf{Average Accuracy (\%)}} \\
\hline
iDT+HSV \cite{c46} 										& \multicolumn{1}{c|}{87.9} \\
\hline
MoFAP \cite{c45} 										& \multicolumn{1}{c|}{88.3} \\
\hline
Two-Stream CNN \cite{c12} 								& \multicolumn{1}{c|}{88.0} \\
\hline
C3D (3 nets) \cite{c13} 								& \multicolumn{1}{c|}{85.2} \\
\hline
C3D (3 nets)+iDT \cite{c13} 							& \multicolumn{1}{c|}{90.4} \\
\hline
F\textsubscript{ST}CN (SCI Fusion) \cite{c14} 			& \multicolumn{1}{c|}{88.1} \\
\hline
TDD+FV \cite{c47} 										& \multicolumn{1}{c|}{90.3} \\
\hline
KVMF \cite{c48} 										& \multicolumn{1}{c|}{93.1} \\
\hline\hline
\textbf{DDM}										    & \multicolumn{1}{c|}{\textbf{91.5}} \\
\hline
\textbf{DDM+SCSP} 									    & \multicolumn{1}{c|}{\textbf{94.3}} \\
\hline
\end{tabularx}
\label{Table1}}
\vspace{-2.3em}
\end{table}

\textit{The Hollywood II dataset} \cite{c36} has been constructed from 69 different Hollywood movies and includes 12 activity classes. It contains a total of 1,707 videos with 823 training videos and 884 testing videos. The length of videos varies from hundreds to several thousand frames. According to the test protocol, the performance is measured by the mean average precession over all classes \cite{c36}.

To compare our approach with the benchmark, we obtain the average precession performance for each class and take the mean average precession (mAP) as indicated in Table \ref{Table2}. The best result is obtained using DDM with a 0.4 mAP improvement in the overall accuracy. The superior performance of the proposed method in action recognition task demonstrates the effectiveness of our long-term spatiotemporal modeling of videos approach.
\begin{table}
\vspace{-2.5em}
\centering
\scriptsize{
\caption{Comparison of the mean average precession (mAP) of DDM with the state-of-the-art methods on the Hollywood II dataset \cite{c36}.}
\begin{tabularx}{\columnwidth}{@{\extracolsep{\fill}}|p{0.5\columnwidth}p{0.5\columnwidth}|}
\hline
\multicolumn{1}{|c}{\textbf{Method}} & \multicolumn{1}{c|}{\textbf{mAP (\%)}} \\
\hline
DL-SFA \cite{c49} 										& \multicolumn{1}{c|}{48.1} \\
\hline
iDT \cite{c44} 											& \multicolumn{1}{c|}{64.3} \\
\hline
Actons \cite{c50} 										& \multicolumn{1}{c|}{64.3} \\
\hline
MIFS \cite{c51} 										& \multicolumn{1}{c|}{68.0} \\
\hline
NL-RFDRP+CNN \cite{c52} 								& \multicolumn{1}{c|}{70.1} \\
\hline
HRP \cite{c14} 											& \multicolumn{1}{c|}{76.7} \\
\hline\hline
\textbf{DDM} 											& \multicolumn{1}{c|}{\textbf{75.3}} \\
\hline
\textbf{DDM+SCSP} 										& \multicolumn{1}{c|}{\textbf{77.1}} \\
\hline
\end{tabularx}
\label{Table2}}
\vspace{-10ex}
\end{table}

\subsection{Dynamic Texture and Dynamic Scene Recognition}

We evaluated the capability of our proposed method in the case of dynamic texture and dynamic scene classification using DynTex \cite{c16} dataset, and YUPENN \cite{c37} and Maryland \cite{c38} datasets, respectively. In order to follow the standard comparison protocol, we use Leave-One-Out (LOO) cross validation.  Note that the results are drawn from the related papers.

\textit{The DynTex dataset} \cite{c16} is a standard database for dynamic texture analysis containing high-quality dynamic texture videos such as windmill, waterfall, and sea waves. It includes over 650 videos recorded in PAL format in various conditions.  Each video has 250 frames length with a 25 frames per second frame rate. Table \ref{Table3} compares the rank-1 recognition rates of DDM with the benchmark approaches. It can be clearly seen that our proposed approach yields in the best results compared to all other methods. 
\begin{table}
\vspace{-2.2em}
\centering
\scriptsize{
\caption{Comparison of the rank-1 recognition rates on the DynTex dataset \cite{c16}.}
\begin{tabularx}{\columnwidth}{@{\extracolsep{\fill}}|p{0.45\columnwidth}p{0.5\columnwidth}|}
\hline
\multicolumn{1}{|c}{\textbf{Method}} & \multicolumn{1}{c|}{\textbf{Recognition Rate (\%)}} \\
\hline
VLBP \cite{c18} 										& \multicolumn{1}{c|}{95.71} \\
\hline
LBP-TOP \cite{c18} 										& \multicolumn{1}{c|}{97.14} \\
\hline
DFS \cite{c39} 											& \multicolumn{1}{c|}{97.63} \\
\hline
BoS Tree \cite{c42} 									& \multicolumn{1}{c|}{98.86} \\
\hline
MBSIF-TOP \cite{c43} 									& \multicolumn{1}{c|}{98.61} \\
\hline
st-TCoF \cite{c9} 										& \multicolumn{1}{c|}{98.20} \\
\hline\hline
\textbf{DDM} 											& \multicolumn{1}{c|}{\textbf{98.05}} \\
\hline
\textbf{DDM+SCSP} 										& \multicolumn{1}{c|}{\textbf{99.27}} \\
\hline
\end{tabularx}
\label{Table3}}
\vspace{-2.2em}
\end{table}

\textit{The YUPENN dataset} \cite{c37} is a stabilized dynamic scene dataset. This dataset was created with an emphasis on scene-specific temporal information. YUPENN contains 14 dynamic scene categories with 30 videos per category. There are significant variations in this dataset's video sequences such as frame rate, scene appearance, scaling, illumination, and camera viewpoint. We report the experimental results on this dataset in Table \ref{Table4}. It can be observed from Table \ref{Table4} that DDM outperforms the existing state-of-the-art methods in the case of dynamic scene classification. The results confirm that the proposed DDM is effective for dynamic scene data classification in a stabilized setting.
\begin{table}
\vspace{-2.2em}
\centering
\scriptsize{
\caption{Comparison of the classification results (\%) on the YUPENN \cite{c37} and Maryland \cite{c38} dynamic scene datasets.}
\begin{tabularx}{\columnwidth}{@{\extracolsep{\fill}}|p{0.4\columnwidth}p{0.3\columnwidth}p{0.3\columnwidth}|}
\hline
\multicolumn{1}{|c}{\textbf{Method}} & \multicolumn{1}{c}{\textbf{YUPENN}} & \multicolumn{1}{c|}{\textbf{Maryland}} \\
\hline
CSO \cite{c40}	       & \multicolumn{1}{c}{85.95} 			& \multicolumn{1}{c|}{67.69} \\
\hline
SFA \cite{c41}	       & \multicolumn{1}{c}{85.48}			& \multicolumn{1}{c|}{60.00} \\
\hline
SOE \cite{c3}	       & \multicolumn{1}{c}{80.71}			& \multicolumn{1}{c|}{43.10} \\
\hline
BoSE \cite{c3} 	       & \multicolumn{1}{c}{96.19}			& \multicolumn{1}{c|}{77.69} \\
\hline
LBP-TOP \cite{c18}     & \multicolumn{1}{c}{84.29}			& \multicolumn{1}{c|}{39.23} \\
\hline
C3D \cite{c13}		   & \multicolumn{1}{c}{98.10}			& \multicolumn{1}{c|}{N/A} \\
\hline
st-TCoF \cite{c9}	   & \multicolumn{1}{c}{99.05}		    & \multicolumn{1}{c|}{88.46} \\
\hline\hline
\textbf{DDM}  		   & \multicolumn{1}{c}{\textbf{97.52}} & \multicolumn{1}{c|}{\textbf{86.33}} \\
\hline
\textbf{DDM+SCSP}      & \multicolumn{1}{c}{\textbf{99.18}} & \multicolumn{1}{c|}{\textbf{90.27}} \\
\hline
\end{tabularx}
\label{Table4}}
\vspace{-2.2em}
\end{table}

\textit{The Maryland dataset} \cite{c38} is a dynamic scene database which consist of 13 natural scene categories containing 10 videos each with 617 frames on average. The dataset has videos showing a wide range of variations in natural dynamic scenes, e.g. avalanches, traffic, and forest fire. One notable difference between the Maryland dataset and the YUPENN dataset is that the former includes camera motions, while the latter does not.

We present the comparison between our proposed method and the state-of-the-art methods in Table \ref{Table4}. Since most of the videos in the Maryland dataset show significant temporal variations, the experimental results suggest that, for highly dynamic data, DDM is able to outperform its strongest rival st-TCoF by a margin of 1.81\%. The promising performance of st-TCoF in the dynamic scene classification (Table \ref{Table4}) is due to incorporating the spatial and the temporal information of the video sequence. However, the results on Maryland dataset suggests that st-TCoF is sensitive to the significant camera motions. On the other hand, our DDM is strongly robust when the structure of the images drastically changes their position with time. Therefore, the DDM can effectively learn the complex underlying structure of the dynamic scene in the presence of severe camera movements.

\vspace{-0.5em}
\subsection{Discriminability Analysis}
\begin{figure}[t]
\vspace{-1.7em}
\centering
\includegraphics[width=\linewidth]{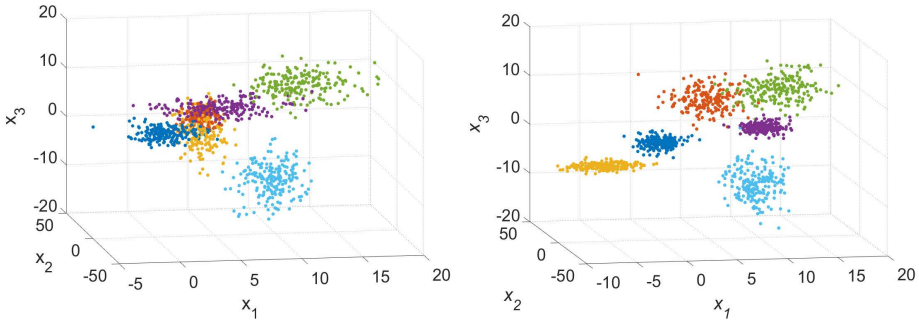}
\caption{An example of the distribution of the learnt classes from UCF101 dataset \cite{c35} before (\textbf{Left}) and after (\textbf{Right}) applying the proposed SCSP. The SCSP reduces the correlation between similar classes by condensing and scattering their samples in the feature space.}
\label{Distribution}
\vspace{-1.5em}
\end{figure}

\vspace{-0.1em}
In order to illustrate the disciminability power of the SCSP, Figure \ref{Distribution} shows the distribution of the sampled data from different classes of UCF101 database before and after applying SCSP in a 3D space. Thanks to the existing redundancy, the samples are correlated before applying SCSP, which makes the data reconstruction a non-trivial task. However, the samples become scattered in the feature space after applying SCSP, i.e. the similar samples are closer to each other and dissimilar samples are far apart. This strategy makes the process of learning class-specific models easier for DDM by learning the underlying structure of each class from the feature space instead of raw video data. 

By enlarging the inter-class similarity of data, the proposed DDM reconstructs the videos of each class more effectively by learning the class-specif models. According to the distances between samples in the new feature space, the DDM can easily learn the pattern and the structure of each class, since the correlation and the redundancy are lessened by applying SCSP.
\vspace{-2ex}
\section{Conclusion}
\vspace{-1.5ex}
We proposed a novel deep learning approach for video-based scene classification. Specifically, a multi-layer deep autoencoder structure was presented which is first pre-trained for appropriate parameter initialization and then fine-tuned for learning class-specific Deep Discriminative Models (DDMs). Capturing the underlying non-linear complex geometric surfaces, the DDMs can effectively model the spatiotemporal variations within video sequences. In order to discard the redundant information in video sequences and avoid the strong correlations between adjacent frames, we captured the spatiotemporal variations in the video sequences and represented them sparsely using Sparse Cubic Symmetrical Pattern (SCSP). The learnt DDMs are used for a minimum reconstruction error-based classification technique during the testing phase. The proposed method has been extensively evaluated on a number of benchmark video datasets for action recognition, dynamic texture and dynamic scene classification tasks. Comparisons against state-of-the-art approaches showed that our proposed method achieves very interesting performance.

\bibliographystyle{splncs}
\bibliography{egbib}
\end{document}